\documentclass{amsart}
\usepackage[utf8]{inputenc}
\usepackage{amsmath,amssymb,amsthm}
\usepackage{tikz}
\usepackage{graphicx}
\usepackage{xcolor}
\usepackage[hidelinks]{hyperref}
\usepackage[ruled,vlined,linesnumbered]{algorithm2e}

\usepackage{xspace}
\usepackage{float}
\usepackage[section]{placeins}  

\title[Accelerating MCTS with Optimized Policies]{Accelerating Monte--Carlo Tree Search with Optimized Posterior Policies}

\author{Keith Frankston}
\address{Center for Communications Research, Institute for Defense Analyses\\
Princeton, NJ 08540, USA}
\email{k.frankston@fastmail.com}
\email{k.frankston@idaccr.org}

\author{Benjamin Howard}
\address{Center for Communications Research, Institute for Defense Analyses\\
Princeton, NJ 08540, USA}
\email{bhoward73@gmail.com}
\email{bjhowa3@idaccr.org}

\subjclass[2020]{cs.AI, cs.LG}


\DeclareMathOperator{\ucb}{ucb}
\DeclareMathOperator{\sgn}{sgn}
\DeclareMathOperator{\KLDiv}{KL-Div}

\newcommand{\RMCTS}{\textsc{RMCTS}\xspace}
\newcommand{\MCTS}{\textsc{MCTS-UCB}\xspace}

\begin{document}

\begin{abstract}
We introduce a recursive AlphaZero-style Monte--Carlo tree search algorithm, ``\RMCTS''.
The advantage of \RMCTS over AlphaZero's \MCTS \cite{silver2018general} is speed. 
In \RMCTS, the search tree is explored in a breadth-first manner, so that 
network inferences naturally occur in large batches.  This significantly reduces 
the GPU latency cost. 
We find that \RMCTS is often more than 40 times faster than \MCTS when searching a single root state, 
and about 3 times faster when searching a large batch of root states.

The recursion in \RMCTS is based on computing optimized posterior policies at each game state in the search tree, 
starting from the leaves and working back up to the root.
Here we use the posterior policy explored in ``Monte--Carlo tree search as regularized policy optimization'' \cite{grill2020}.
Their posterior policy is the unique policy which maximizes 
the expected reward given estimated action rewards  
minus a penalty for diverging from the prior policy.

The tree explored by \RMCTS is not defined in an adaptive manner, as it is in \MCTS.  Instead, the \RMCTS tree is defined by following prior network policies at each node.
This is a disadvantage, but the speedup advantage is more significant, and in practice we 
find that \RMCTS-trained networks match the quality of \MCTS-trained networks in roughly one-third of the 
training time. We include timing and quality comparisons of \RMCTS vs. \MCTS for three games: Connect--4, Dots-and-Boxes, and Othello.

\end{abstract}

\maketitle

\section{AlphaZero's \MCTS}

AlphaZero \cite{silver2018general} is a method to train neural networks to play games at a high level.
The main idea is to use Monte-Carlo tree search (MCTS) to explore the game tree and
learn an improved value and policy for a given game state, where the exploration is
based on prior values and policies from the current network.
The network is trained on these improved values and policies, and becomes stronger over time.

In AlphaZero, Monte-Carlo tree search (\MCTS) 
works roughly as follows (see \autoref{alg:mcts} in \autoref{sec:algorithms} for a more precise description). 
We initialize estimated action values $Q(s,\cdot)$ to zero, and then 
perform a number of simulations to refine the $Q$ values and obtain a posterior policy.
A simulation always begins at the root state, 
and chooses an action with the maximal UCB value, where the UCB value is defined to be 
\[ \ucb(s,a) = Q(s,a) + C \cdot \pi_0(s,a) \cdot \frac{\sqrt{\sum_{b} N(s,b)}}{1 + N(s,a)} \]
where $s$ is the current state, $a$ is an available action from $s$, 
$Q(s,a)$ is the estimated reward for taking action $a$ from state $s$,
$C > 0$ is the exploration constant,  
$N(s,a)$ is the number of times action $a$ has been taken from state $s$ in previous simulations, 
and finally, $\pi_0(s,\cdot)$ is the prior (network) policy.

We continue picking the action with highest UCB, moving down the exploration tree, until we reach a  
state $s'$ that is terminal, or which is not yet in the tree. 
If $s'$ is terminal, then its value is simply the game score.
Otherwise, the value of $s'$ is defined to be the network value $v_0(s)$, and $s'$ is 
added to the tree, initializing $Q(s',\cdot)$ and $N(s',\cdot)$ to zero for all actions from $s'$.
The value of this final state is propagated back up the path to the root using the appropriate 
sign for the active player; each state along 
this path updates its $Q$ and $N$ values appropriately.

Updating the $Q$ and $N$ values causes the UCB values to change, and so 
a different path may be taken in the next simulation. 
Once $N(s) = \sum_a N(s,a)$ reaches the budgeted number of simulations, the process stops.
At this point,     
the posterior policy $\hat\pi$ at a state $s$ is defined to be the normalized visit counts 
$\hat\pi(s,a) = \frac{N(s,a)}{N(s)}$.

\begin{figure}[h!]
\centering
\includegraphics[scale=0.5]{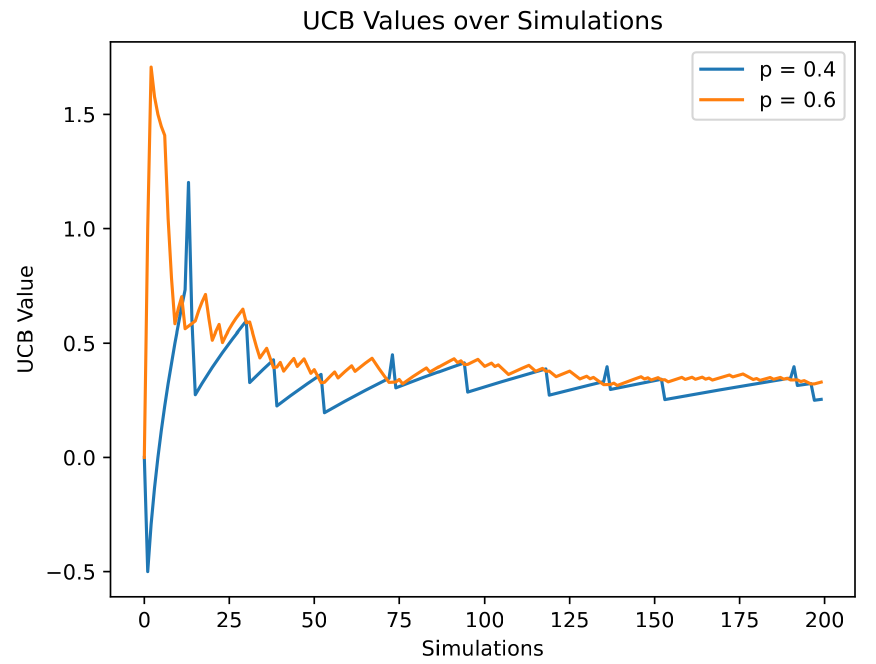}
\caption{Illustration of UCB values over 200 simulations in bandit game with two slot arms.}\label{fig:bandit}
\end{figure}

\autoref{fig:bandit} illustrates the evolution of UCB values for a simple (one-player) bandit game with two slot arms, 
starting from a uniform prior policy.
Each slot pays out a reward of $+1$ or $-1$; the probability of $+1$ is a fixed 
hidden parameter $p$.  For one slot, $p = 0.6$, and for the other, $p = 0.4$.
The inferior choice ($p = 0.4$) is never abandoned; this action 
is chosen $\Theta(1/\sqrt{N})$ of the time, where $N$ is the total number of simulations. 
As $N \to \infty$, the two UCB values get closer and closer together, approaching $0.2 + \Theta(1 / \sqrt{N})$.

\section{Regularized Policy Optimization}

An alternative to $\hat\pi$ was explored in ``Monte-Carlo tree search as 
regularized policy optimization'' \cite{grill2020}.
They define the optimized posterior policy $\bar{\pi}(s,\cdot)$ to be the one maximizing
the expected reward (assuming estimated rewards $Q(s,a)$) minus a penalty for diverging from the prior policy 
$\pi_0(s, \cdot)$.
Specifically, $\bar\pi(s,\cdot)$ is the unique distribution on actions at game state $s$ which maximizes
\[ \sum_a \bar\pi(s,a) Q(s,a) - \frac{C}{\sqrt{N(s)}} \KLDiv(\pi_0(s,\cdot) \mid\mid \bar\pi(s, \cdot)) \]
where $\KLDiv$ is the Kullback-Leibler divergence, and $C > 0$ is the same exploration constant 
as used in \MCTS.
They show that $\bar{\pi}$ agrees asymptotically to $\hat{\pi}$ (normalized visit counts) as the number of MCTS simulations approaches infinity,
but argue that $\bar{\pi}$ is superior to $\hat{\pi}$ when the number of simulations is small.
They suggest replacing $\hat{\pi}$ with $\bar{\pi}$ in the AlphaZero algorithm, where the $Q$-values are computed using the original
\MCTS variant of AlphaZero.

We note that the optimized posterior policy $\bar{\pi}$ has an interesting interpretation:  
it is essentially the policy which forces the UCB values of all actions to be equal.
Recall that the UCB value is defined as 
$$\ucb(s,a) = Q(s,a) + C \cdot \pi_0(s,a) \cdot \frac{\sqrt{N(s)}}{1 + N(s,a)}.$$
Now suppose that we have finished all the simulations at state $s$, and 
$\hat{\pi}(s,a) = \frac{N(s,a)}{N(s)}$ is the normalized visit counts.
For simplicity, let us drop the ``$1 + $'' in the denominator of the exploration term.
Now we can rewrite the UCB value as
\begin{align*}
\ucb(s,a) &= Q(s,a) + C \cdot \pi_0(s,a) \cdot \frac{\sqrt{N(s)}}{\hat{\pi}(s,a) N(s)} \\
&= Q(s,a) + \frac{C}{\sqrt{N(s)}} \cdot \frac{\pi_0(s,a)}{\hat{\pi}(s,a)}.
\end{align*}
Now replace $\hat{\pi}(s,a)$ with a variable $\bar{\pi}(s,a)$, and consider the problem of maximizing
the objective function 
$$F(\bar{\pi}(s,\cdot)) = \sum_a \bar{\pi}(s,a) Q(s,a) - \frac{C}{\sqrt{N(s)}} \KLDiv(\pi_0(s,\cdot) \mid\mid \bar{\pi}(s, \cdot)),$$ 
subject to the constraint that $\sum_a \bar{\pi}(s,a) = 1$.
A simple calculation shows that $\frac{\partial F}{\partial \bar{\pi}(s,a)} = \ucb(s,a)$.
On the other hand, the constraint function \(\sum_a \bar{\pi}(s,a) = 1\) has partial derivative \(1\) with respect to each variable \(\bar{\pi}(s,a)\).
Hence the method of Lagrange multipliers tells us that the optimum \(\bar{\pi}(s,\cdot)\) can
only occur where all the UCB values \(\ucb(s,a)\) are equal.
From \autoref{fig:bandit} one can see that the UCB values 
of \MCTS approach each other over time; this optimized posterior policy $\bar{\pi}$ forces them to be equal.

If $u$ is the common UCB value for the optimal posterior policy $\bar{\pi}(s,\cdot)$, then we have
$$\bar{\pi}(s,a) = \frac{C}{\sqrt{N(s)}} \frac{\pi_0(s,a)}{u - Q(s,a)}.$$
In particular, we are looking for the unique value of $u > \max_a Q(s,a)$ such that 
$$\sum_a \frac{C}{\sqrt{N(s)}} \frac{\pi_0(s,a)}{u - Q(s,a)} = 1.$$
This value certainly exists, since as $u \to \max_a Q(s,a)^+$, the left-hand side approaches $+\infty$,
whereas as $u \to +\infty$, the left-hand side approaches $0$.

The optimal posterior policy $\bar{\pi}$ can be computed efficiently using Newton's method (cf. \autoref{alg:eucb} in \autoref{sec:algorithms}). 
    In this algorithm, the function $f$ is convex and since we begin with the initial value on the appropriate side of the
    solution, Newton's method is guaranteed to converge.

\section{\RMCTS}
An important feature of the optimized posterior policy $\bar{\pi}$ is that it can be computed 
locally at each game state $s$, using only the estimated action values $Q(s,\cdot)$ and the prior policy $\pi_0(s,\cdot)$.
It does not require any details about the rest of the search tree.  This is what permits us 
to define a recursive alternative to \MCTS, which we call \RMCTS.

Recall that \cite{grill2020} suggests replacing the posterior policy $\hat{\pi}$ with the optimized 
posterior policy $\bar{\pi}$, where the estimated $Q$ values are gotten from the original \MCTS variant of AlphaZero.
By contrast, \RMCTS completely redefines \mbox{MCTS} itself, 
by using optimized policies not only at the root state, but throughout the tree.     
The $Q$-values are computed in a recursive manner, where the value of
    nodes $s'$ in the search tree are approximated (recursively) by computing optimized posterior policies below $s'$. 
See \autoref{alg:rmcts} in \autoref{sec:algorithms} for a succinct description of \RMCTS.

The search tree is generated by following prior network policies at each node (in particular it is
    not defined by UCB values).  
    Each node $s$ in the tree consumes one simulation\footnote{Since GPU inference is the bottleneck, 
    and we need such an inference at every node of the tree, 
    we define the simulation count to drop by one at every node.}, and then awards the 
    remaining simulations to its child actions according to the prior policy $\pi_0(s,\cdot)$.
    Thus if a state $s$ is afforded $N(s)$ simulations, then one simulation is used to 
    acquire the prior policy $\pi_0(s,\cdot)$ and value $v_0(s)$ from the network,
    leaving $N(s)-1$ simulations to be distributed among the available actions from $s$.
    The number of simulations $N(s,a)$ assigned to action $a$ (cf. \autoref{alg:assign_sims}) 
    has expectation $\mathbb{E}[N(s,a)] = \pi_0(s,a) (N(s)-1)$, where $\pi_0$ is 
    the prior network policy at state $s$.    
    As in \MCTS, if $s'$ is a nonterminal leaf (afforded one simulation), 
    then its value is defined to be the network value $v_0(s')$.
    If $s'$ is a terminal state, then its value is simply the game score.
    See \autoref{sec:simple_example} for a simple example illustrating \RMCTS.

\subsection{Implementing \RMCTS efficiently}    
The description of \RMCTS in \autoref{sec:algorithms} is mathematically precise, but it is 
not efficient to implement this way, where the function calls itself recursively.
See \url{https://github.com/bhoward73/rmcts} for an efficient C implementation of \RMCTS.
This efficient implementation of \autoref{alg:rmcts} works iteratively, exploring  
the search tree in a breadth-first manner.
All nodes at the same depth form one large batch of GPU inferences, 
and so the GPU latency cost is largely mitigated.
Contiguous memory is pre-allocated for all the nodes (and relevant data) for the search tree, 
improving cache performance.

\subsection{Timings and quality comparisons}
The timings we report in \autoref{sec:timings} are based on the efficient implementation 
of \RMCTS described above.
We compare the quality of \RMCTS and \MCTS in \autoref{sec:quality_comparison}, 
by pitting them against each other in three games: Connect-4, Dots-and-Boxes, and Othello.
In this contest, both \RMCTS and \MCTS use the same neural network for priors.

\section{Algorithm Descriptions}\label{sec:algorithms}

In this section we give precise descriptions of \MCTS (\autoref{alg:mcts}) and \RMCTS (\autoref{alg:rmcts}).

\begin{algorithm}[H]
\caption{\MCTS}
\label{alg:mcts}
\KwIn{State $s$, number of simulations $N$, exploration constant $C$}
\KwOut{New policy $\hat\pi$ at root state $s$}
Given state $t$, let $\sgn(t) = +1$ if player 1 to move in state $t$, else $-1$\;
\mbox{visited} = $\emptyset$\;
\While{$N > 0$}{
    $t \gets s$\;
    \mbox{path} = []\;
    \While{$t$ is in \mbox{visited} and $t$ is nonterminal}{
        Select action $a$ that maximizes $\ucb(t,a) = Q(t,a) + C \cdot \pi_0(t,a) \frac{\sqrt{\sum_b N(t,b)}}{1 + N(t,a)}$\;
        Append $(t,a)$ to \mbox{path}\;
        $t \gets$ state reached by taking action $a$ from state $t$\;
    }
    \eIf{$t$ is nonterminal}{
        Add $t$ to \mbox{visited}\;
        $Q(t,a) \gets 0$ for all actions $a$\;
        $N(t,a) \gets 0$ for all actions $a$\;
        Acquire priors $v_0(t)$ and $\pi_0(t, \cdot)$ from neural network\;
        $v \gets \sgn(t) v_0(t)$ (appropriate sign relative to player 1)\;
    }{
        $v \gets$ score of terminal state $t$, relative to player 1\;
    }
    \ForEach{$(t,a)$ in \mbox{path}}{
        $N(t,a) \gets N(t,a) + 1$\;
        $Q(t,a) \gets Q(t,a) + \frac{\sgn(t) v - Q(t,a)}{N(t,a)}$\;
    }
    $N \gets N - 1$\;
}
\ForEach{action $a$}{
    $\hat\pi(a) \gets \frac{N(s,a)}{\sum_{b} N(s,b)}$\;
}
\Return{$\hat\pi$}
\end{algorithm}

\begin{algorithm}[H]
\caption{\RMCTS (deterministic, two-player zero-sum)}
\label{alg:rmcts}
\KwIn{State $s$, number of simulations $N$, exploration constant $C$}
\KwOut{Estimated value $\bar{v}$ and policy $\bar\pi$ at root state $s$}
\If{$s$ is terminal}{
    \Return{score of $s$, NULL policy}\;
}
Let $\sgn(s,t) = +1$ if same player is active for states $s$ and $t$, else $-1$\;
Acquire priors $v_0(s)$ and $\pi_0(s, \cdot)$ from neural network\;
$N(s,\cdot) \gets \textsc{ASSIGN-SIMULATIONS}(s, N-1, \pi_0(s,\cdot))$; \tcp{\autoref{alg:assign_sims}}
$A \gets \{ a : N(s,a) > 0 \}$\;
\ForEach{action $a \notin A$}{
    $\bar{\pi}(s,a) \gets 0$\; 
}
\ForEach{action $a \in A$}{
    let $t$ be the state reached by taking action $a$ from state $s$\;
    $v_t, \pi_t \gets \textsc{\RMCTS}(t, N(s,a), C)$\;
    $Q(s,a) \gets \sgn(s,t) \, v_t$\;
}
Let $Q', \pi'_0$ be the restriction of $Q(s,\cdot)$ and $\pi_0(s,\cdot)$ to actions in $A$\;
$\pi' \gets \mbox{POLICY-OPTIMIZATION}(Q', \pi'_0, N-1, C)$; \tcp{\autoref{alg:eucb}}
\ForEach{action $a \in A$}{
    $\bar{\pi}(s,a) \gets \pi'(a)$\;
}
$\bar{v} \gets \frac{1}{N} \cdot v_0(s) + \frac{N-1}{N} \cdot \sum_{a \in A} Q(s,a) \cdot \bar\pi(s,a)$\;
\Return{$\bar{v}, \bar\pi$}
\end{algorithm}

\begin{algorithm}[H]
\caption{ASSIGN-SIMULATIONS}
\label{alg:assign_sims}
\KwIn{State $s$, number of simulations $N$, prior policy $\pi_0(s,\cdot)$}
\KwOut{Number of simulations $N(s,a)$ assigned to each action $a$}
Put all potential actions in some arbitrary order $a_1, a_2, \ldots, a_{n}$\;
$t_0 \gets 0$\;
\For{$i = 1, 2, \ldots, n$}{
    $t_i \gets N \; \sum_{j \leq i} \pi_0(s, a_j)$\;
}
Generate a uniformly random number $x$ in $[0,1)$\;
\For{$i = 1, 2, \ldots, n$}{
    $N(s, a_i) \gets \# \{ k \in \mathbb{Z} : t_{i-1} \leq x + k < t_i \}$\;
}
\Return{$N(s,a)$ for each action $a$}
\end{algorithm}

The subroutine \autoref{alg:assign_sims} (ASSIGN-SIMULATIONS) randomly distributes a total of $N$ simulations to actions according to the 
prior policy $\pi_0(s,\cdot)$.  Action $a$ is assigned $N(s,a)$ simulations.
The expectation $\mathbb{E}[N(s,a)] = \pi_0(s,a) N$, 
and $\lfloor \pi_0(s,a) N \rfloor \leq N(s,a) \leq \lceil \pi_0(s,a) N \rceil$.
We note that, in particular, every action $a$ is guaranteed to receive at least $\lfloor \pi_0(s,a) N \rfloor$ simulations.
Finally, we have $\sum_a N(s,a) = N$.

\begin{algorithm}[H]
\caption{POLICY-OPTIMIZATION}
\label{alg:eucb}
\KwIn{estimated action rewards $Q$, prior policy $\pi_0$, number of simulations $N$, exploration constant $C$}
\KwOut{Updated policy $\bar\pi$ as defined in \cite{grill2020}, maximizing $\sum_a \bar\pi(a) Q(a) - \frac{C}{\sqrt{N}} \KLDiv(\pi_0 \mid\mid \bar\pi)$}
$\epsilon \gets 10^{-10}$\;
$\lambda \gets C / \sqrt{N}$\;
Define $f(u) \gets -1 + \lambda \sum_a \frac{\pi_0(a)}{u - Q(a)}$\;
$u \gets \max_a Q(a) + \lambda \pi_0(a)$\;
\While{$f(u) > \epsilon$}{
    $u \gets u - \frac{f(u)}{f'(u)}$ \tcp{Newton's method always converges here}
}
\ForEach{action $a$}{
    $\bar\pi(a) \gets \lambda \frac{\pi_0(a)}{u - Q(a)}$\;
}
Normalize $\bar\pi$ so that $\sum_a \bar\pi(a) = 1$\;
\Return{$\bar\pi$}
\end{algorithm}

\section{A simple example}\label{sec:simple_example}
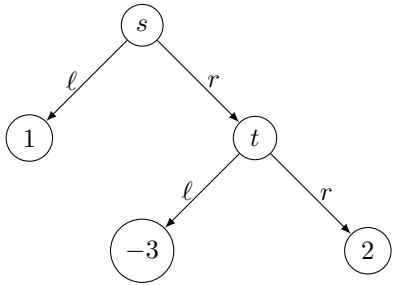
\begin{figure}[h!]
\centering
\begin{tikzpicture}[level distance=1.5cm, sibling distance=3cm, edge from parent/.style={draw,-latex}]
    \node[circle, draw] (s) {$s$}
        child {
            node[circle, draw] (x) {$1$}
            edge from parent node[left] {$\ell$}
        }
        child {
            node[circle, draw] (t) {$t$}
            child {
                node[circle, draw] (y) {$-3$}
                edge from parent node[left] {$\ell$}
            }
            child {
                node[circle, draw] (z) {$2$}
                edge from parent node[right] {$r$}
            }
            edge from parent node[right] {$r$}
        };
\end{tikzpicture}
\caption{Clearly 
the best action from state $s$ in this one-player game is to go right, reaching 
state $t$, where we should again choose to go right, obtaining the maximal reward of $2$.}
\label{fig:binary_tree}
\end{figure}

Consider the following toy-sized one-player game illustrated in \autoref{fig:binary_tree}, where
the game tree is a binary tree with only two nonterminal states $s$ and $t$.
The available actions are $\ell$ (left) and $r$ (right).  
Starting from the root state $s$, if we go left, we reach a terminal state 
with value $1$.  If instead we go right, we reach the nonterminal state $t$.
From $t$, if we go left, we reach a terminal state with value $-3$.
If we go right from $t$, we reach a terminal state with value $2$.

Suppose that we are afforded $N = 1003$ simulations at the root state $s$.
Initially we have no strong opinions and our prior policy is uniform, and our prior value is zero 
on all nonterminal states. Let's set our exploration constant to $C = 1$.

We spend one simulation on $s$ itself, for which our prior value is $v_0(s) = 0$, leaving us 
with $1002$ simulations to distribute among the two actions $\ell$ and $r$ from $s$.
Since the prior policy is uniform we assign $N(s, \ell) = 501$ simulations to action $\ell$ and also 
$N(s, r) = 501$ simulations to action $r$.
Since the left action $\ell$ from $s$ reaches a terminal state with value $1$, we know the 
exact value $Q(s,\ell) = 1$ for the left action.
For the right action $r$ from $s$, we first consume one simulation to obtain 
the uniform prior policy and zero prior value at state $t$.
We assign $N(t, \ell) = 250$ simulations to action $\ell$ and
$N(t, r) = 250$ simulations to action $r$ from state $t$.

Now that the search tree and simulation counts have been defined, 
we now turn to the recursive computation of $Q$ values. 
The optimized posterior policy at state $t$ (\autoref{alg:eucb}) is 
approximately $\bar{\pi}(t, \ell) = 0.00445$ and $\bar{\pi}(t, r) = 0.996$.
Giving one vote to the prior value $v_0(t) = 0$, our new estimated value $\bar{v}(t) = Q(s,r)$ is 
\[ Q(s,r) = \bar{v}(t) = \frac{1}{501} \cdot 0 + \frac{500}{501} \cdot (-3 \cdot 0.00445 + 2 \cdot 0.996) \approx 1.98.\]
Now that all estimated $Q$ values at state $s$ are known, we can compute the optimized posterior policy at $s$.
The optimized posterior policy at the root state $s$ is computed (\autoref{alg:eucb}) to be approximately $\bar{\pi}(s, \ell) = 0.016$ and $\bar{\pi}(s, r) = 0.984$.
Thus, our final estimated value at the root state $s$ is 
\[\bar{v}(s) = \frac{1}{1003} \cdot 0 + \frac{1002}{1003} \cdot (1 \cdot 0.016 + 1.98 \cdot 0.984) \approx 1.96.\]
Hence the posterior value of $s$ is approximately $1.96$, and the posterior policy at $s$
strongly favors going right with a probability of $98.4 \%$.

Note that it was very important to use recursion to define the $Q$ values.  If we had assigned 
the $Q$ values to be the expected action values following the prior policy $\pi_0$ (no recursion), then 
our new posterior policy $\bar{\pi}$ would have favored going left from $s$, since 
going left yields a reward of $1$, whereas the expected value of going right is 
$(-3 + 2) / 2 = -0.5 < 1$.  

\section{Timings for \MCTS vs \RMCTS for three games}\label{sec:timings}
In this section we give timing comparisons of \MCTS vs \RMCTS for three games: Connect-4, Dots-and-Boxes, and Othello.
All timings use TensorRT to optimize the neural network inferences.
We ran these tests on a desktop computer which has an NVIDIA RTX 3080 GPU, and an Intel i7-10700K CPU.
In the cases of Dots-and-Boxes and Othello, the network was a ResNet with 8 residual blocks and 48 channels.
For Connect-4, the network was a ResNet with 8 residual blocks and 64 channels.
In all cases the kernel size was $3 \times 3$.

We first consider the case where we are timing MCTS for a single root state.  
This situation applies when we (human) play against a pre-trained AI opponent, 
and we want to supplement the AI network with MCTS at each move.
Here \RMCTS has the greatest speed advantage over \MCTS.

\begin{table}[H]
\centering
\begin{tabular}{ |c||ccccccc|} 
 \hline
$N$ & $32$ & $64$ & $128$ & $256$ & $512$ & $1024$ & $2048$ \\ 
 \MCTS & $13$ ms & $25$ ms & $49$ ms & $98$ ms & $200$ ms & $390$ ms & $790$ ms \\ 
 \RMCTS & $1.9$ ms & $2.5$ ms & $3.2$ ms & $4.2$ ms & $5.8$ ms & $8.9$ ms & $14$ ms \\ 
 speedup & $6.8\times$ & $10\times$ & $16\times$ & $24\times$ & $34\times$ & $44\times$ & $57\times$  \\
 \hline
\end{tabular}
\caption{Othello timings, one root state.}\label{tab:othello_1}
\end{table}

\begin{table}[H]
\centering
\begin{tabular}{ |c||ccccccc|} 
 \hline
$N$ & $32$ & $64$ & $128$ & $256$ & $512$ & $1024$ & $2048$ \\ 
 \MCTS & $13$ ms & $25$ ms & $47$ ms & $93$ ms & $190$ ms & $370$ ms & $730$ ms \\ 
 \RMCTS & $3.3$ ms & $4.7$ ms & $5.7$ ms & $6.7$ ms & $7.3$ ms & $8.0$ ms & $10$ ms \\ 
 speedup & $3.9\times$ & $5.2\times$ & $8.5\times$ & $15\times$ & $26\times$ & $48\times$ & $77\times$  \\
 \hline
\end{tabular}
\caption{Dots-and-Boxes timings, one root state.}\label{tab:dotbox_1}
\end{table}

\begin{table}[H]
\centering
\begin{tabular}{ |c||ccccccc|} 
 \hline
$N$ & $32$ & $64$ & $128$ & $256$ & $512$ & $1024$ & $2048$ \\ 
 \MCTS & $18$ ms & $35$ ms & $70$ ms & $150$ ms & $280$ ms & $550$ ms & $1100$ ms \\ 
 \RMCTS & $3.8$ ms & $4.9$ ms & $6.8$ ms & $9.2$ ms & $9.7$ ms & $12$ ms & $15$ ms \\ 
 speedup & $4.8\times$ & $7.2\times$ & $10\times$ & $16\times$ & $28\times$ & $45\times$ & $74\times$  \\
 \hline
\end{tabular}
\caption{Connect-4 timings, one root state.}\label{tab:connect4_1}
\end{table}

We next consider the case when we are computing MCTS for a batch of root states (trees) ``at once.''
This situation applies when we are using MCTS to generate rollouts for training the neural network.
Here the latency cost of network inferences is largely mitigated for both \MCTS and \RMCTS, 
since the network inferences are done in batches in both cases.
In \MCTS, when searching a batch of root states, we move from one state to next whenever a necessary inference 
prevents us from continuing the computation at the given state.  Once we've passed through all states in the batch, 
then we wait for responses from the network before continuing.  
The latency cost is mitigated in this way, but \RMCTS maintains a significant speed advantage, because 
the batch sizes for \RMCTS are much larger, since 
an \RMCTS batch consists of all nodes at a given depth across all search trees.

\begin{table}[H]
\centering
\begin{tabular}{ |c||ccccccc|} 
 \hline
$N$ & $32$ & $64$ & $128$ & $256$ & $512$ & $1024$ & $2048$ \\ 
 \MCTS & $0.58$ ms & $1.2$ ms & $2.5$ ms & $5.3$ ms & $12$ ms & $27$ ms & $63$ ms \\ 
 \RMCTS & $0.19$ ms & $0.33$ ms & $0.60$ ms & $1.1$ ms & $2.2$ ms & $4.2$ ms & $8.5$ ms \\ 
 speedup & $3.1\times$ & $3.5\times$ & $4.1\times$ & $4.7\times$ & $5.5\times$ & $6.6\times$ & $7.3\times$  \\
 \hline
\end{tabular}

\caption{Othello, 64 root states, average time per root state.}\label{tab:othello_64}
\end{table}

\begin{table}[H]
\centering
\begin{tabular}{ |c||ccccccc|} 
 \hline
$N$ & $32$ & $64$ & $128$ & $256$ & $512$ & $1024$ & $2048$ \\ 
 \MCTS & $0.54$ ms & $1.1$ ms & $2.2$ ms & $4.6$ ms & $9.9$ ms & $23$ ms & $51$ ms \\ 
 \RMCTS & $0.15$ ms & $0.22$ ms & $0.33$ ms & $0.57$ ms & $0.92$ ms & $1.7$ ms & $2.8$ ms \\ 
 speedup & $3.5\times$ & $4.9\times$ & $6.7\times$ & $8.2\times$ & $10.7\times$ & $13.7\times$ & $18.1\times$  \\
 \hline
\end{tabular}

\caption{Dots-and-Boxes, 64 root states, average time per root state.}\label{tab:dotbox_64}
\end{table}

\begin{table}[H]
\centering
\begin{tabular}{ |c||ccccccc|} 
 \hline
$N$ & $32$ & $64$ & $128$ & $256$ & $512$ & $1024$ & $2048$ \\ 
 \MCTS & $0.40$ ms & $0.80$ ms & $1.5$ ms & $3.1$ ms & $6.1$ ms & $12$ ms & $25$ ms \\ 
 \RMCTS & $0.26$ ms & $0.36$ ms & $0.53$ ms & $0.82$ ms & $1.4$ ms & $2.5$ ms & $4.5$ ms \\ 
 speedup & $1.5\times$ & $2.2\times$ & $2.9\times$ & $3.7\times$ & $4.4\times$ & $4.8\times$ & $5.4\times$  \\
 \hline
\end{tabular}

\caption{Connect-4, 64 root states, average time per root state.}\label{tab:connect4_64}
\end{table}


\section{Comparing the quality of \RMCTS vs \MCTS}\label{sec:quality_comparison}

We find that
\RMCTS is at a small disadvantage if the number of simulations is equal.
Most likely the reason for this is that \MCTS creates the search tree in an adaptive manner,
whereas \RMCTS creates the search tree in a non-adaptive manner.
Nevertheless, the speed advantage of \RMCTS more than makes up for this disadvantage. 
See \autoref{tab:othello_scores_table} for an example of this in Othello.
It seems a general rule of thumb that \RMCTS wins games against \MCTS when the number of simulations 
for \RMCTS is twice that of \MCTS, but \RMCTS still takes far less time than \MCTS 
even though it is using twice the number of simulations.

\autoref{tab:othello_scores_table} shows the results of
\RMCTS vs \MCTS in 64 games of Othello (32 played as first player, and 32 as second player).
In both cases, \MCTS had $N = 256$ simulations and required about $2.3$ seconds per game. 
In the first batch of 64 games, \RMCTS also had $N = 256$ simulations, 
but generally lost with a mean score (checker difference) of $-4.0$.
The time required by \RMCTS per game, however, was only $135$ milliseconds.
In the second set of 64 games, \RMCTS was given $N = 512$ simulations, and this time it won with a mean score of $7.2$.
The mean time per game was $178$ milliseconds in this case -- still far less than the \MCTS time of $2.3$ seconds.

\begin{table}[H]
\centering
\begin{tabular}{cccc} \hline
$N$ for \RMCTS & mean score & mean time per game & speedup over \MCTS \\ \hline \hline
$256$ & $-4.0$ & $135$ milliseconds & $17\times$ \\ 
$512$ & $7.2$ & $178$ milliseconds & $13\times$ \\ \hline
\end{tabular}

\caption{Pitting \RMCTS vs \MCTS in 64 games of Othello.  
Doubling the number of simulations for \RMCTS gives it both a score advantage and speed advantage.}\label{tab:othello_scores_table}
\end{table}
  
To compare timings for training with 64 games at once, the average \RMCTS time with $N = 512$ simulations is about $2.2$ milliseconds, 
whereas the average time for \MCTS with $N = 256$ simulations is about $5.3$ milliseconds. 
Hence \RMCTS has the advantage both in quality and time, with a speedup factor of $2.4\times$.
Equating their performance, the speedup increases to above $3\times$.
This is the kind of comparison that matters most for training, and indeed this 
factor of $3\times$ speedup in training time is what we observed in practice.


\section{What next?}
In the near future we plan to test an adaptive variant of \RMCTS.
We do $k$ re-explorations at the same root state, continually refining the policies 
at all explored nodes by taking a weighted average of the prior policy and the posterior policy 
from the most recent exploration.
With each iteration, the subtree goes deeper into branches favored by the posterior policies.
The nodes that were previously explored would not count against the simulation budget 
since we would store their values and policies from the previous explorations.
Thus the overall procedure becomes adaptive, and would likely outperform the non-adaptive version.

We emphasize that \RMCTS is not limited to the specific 
optimal posterior policy of \cite{grill2020} (\autoref{alg:eucb}).  Other variants of optimized posterior policies
can be considered, for example we can replace the Kullback-Leibler divergence with other divergence measures.
It would be interesting to see how well these variants perform.
For example, if we 
reverse the order of $\pi_0$ and $\bar{\pi}$ in the Kullback-Leibler divergence: i.e. $\bar{\pi}$ maximizes 
\[\sum_a \bar\pi(s,a) Q(s,a) - \frac{C}{\sqrt{N(s)}} \KLDiv(\bar\pi \mid\mid \pi_0),\]
then $\bar\pi(s,a) \propto \pi_0(s,a) \exp\left( \frac{\sqrt{N(s)}}{C} Q(s,a) \right)$.
This is an appealing formula, and easy to calculate; but it probably punishes the least-favored actions too much.

In general, \RMCTS makes sense for reasonable variants of \autoref{alg:eucb}, 
where the optimized posterior policy can be computed efficiently, and is computed 
based only on the estimated $Q$-values, prior policy, number of simulations, and exploration constant.

Throughout this paper we have only talked about AlphaZero, and not mentioned its successor MuZero \cite{schrittwieser2020}.
However, \RMCTS also applies to MuZero, since the search procedure of MuZero is once again \MCTS, excepting 
that it is done in the simulated latent space of the learned model.

\section{Acknowledgements}
We thank Tom Sznigir, James Barker, Timothy Chow, Emma Cohen, Ryan Eberhart, Stephen Fischer, Stephen Boyack, Ross Parker, 
and Lawren Smithline (all of IDA/CCR-Princeton) for helpful discussions and comments.
We thank Laurent Sartran and Thomas Hubert of Google/DeepMind for helpful discussions at the Institute for Advanced Study in Princeton, NJ, during the Spring of 2025.
We thank Akshay Venkatesh for inviting the second author to attend the DeepMind presentation at IAS.

\bibliographystyle{plain}
\bibliography{references}

@article{silver2018general,
  title={A general reinforcement learning algorithm that masters chess, shogi, and Go through self-play},
  author={Silver, David and Hubert, Thomas and Schrittwieser, Julian and Antonoglou, Ioannis and Lai, Matthew and Guez, Arthur and Lanctot, Marc and Sifre, Laurent and Kumaran, Dharshan and Graepel, Thore and Lillicrap, Timothy and Simonyan, Karen and Hassabis, Demis},
  journal={Science},
  volume={362},
  number={6419},
  pages={1140--1144},
  year={2018},
  publisher={American Association for the Advancement of Science}
}

@inproceedings{grill2020,
  title={Monte-Carlo tree search as regularized policy optimization},
  author={Grill, Jean-Bastien and Altch{\'e}, Florent and Tang, Yunhao and Hubert, Thomas and Valko, Michal and Antonoglou, Ioannis and Munos, R{\'e}mi},
  booktitle={International Conference on Machine Learning},
  pages={3769--3778},
  year={2020},
  organization={PMLR}
}

@article{schrittwieser2020,
  title={Mastering Atari, Go, chess and shogi by planning with a learned model},
  author={Schrittwieser, Julian and Antonoglou, Ioannis and Hubert, Thomas and Simonyan, Karen and Sifre, Laurent and Schmitt, Simon and Guez, Arthur and Lockhart, Edward and Hassabis, Demis and Graepel, Thore and Lillicrap, Timothy and Silver, David},
  journal={Nature},
  volume={588},
  number={7839},
  pages={604--609},
  year={2020},
  publisher={Nature Publishing Group}
}

\end{document}